\documentclass[runningheads]{llncs}
\usepackage{graphicx}
\usepackage{wrapfig}
\usepackage{upgreek}
\usepackage[dvipsnames]{xcolor}
\usepackage{multirow}
\usepackage{tikz}
\usepackage{comment}
\usepackage{amsmath,amssymb} % define this before the line numbering.
\usepackage{enumerate}
\usepackage{color}

\usepackage[accsupp]{axessibility}  % Improves PDF readability for those with disabilities.

\usepackage[width=122mm,left=12mm,paperwidth=146mm,height=193mm,top=12mm,paperheight=217mm]{geometry}

\newif\ifdraft
\drafttrue

\ifdraft
    \newcommand{\liu}[1]{\textcolor{red}{{[liu: #1]}}}
\else
    \newcommand{\liu}[1]{}
\fi

\begin{document}
% \renewcommand\thelinenumber{\color[rgb]{0.2,0.5,0.8}\normalfont\sffamily\scriptsize\arabic{linenumber}\color[rgb]{0,0,0}}
% \renewcommand\makeLineNumber {\hss\thelinenumber\ \hspace{6mm} \rlap{\hskip\textwidth\ \hspace{6.5mm}\thelinenumber}}
% \linenumbers
\pagestyle{headings}
\mainmatter
\def\ECCVSubNumber{1577}  % Insert your submission number here

\title{
% Diffusion Unit: An Adaptive and Interpretable Edge-Aware Computational Unit for 3D Point Cloud Understanding/Suppress Learning for Interpretable 3D Point Cloud Understanding\\
% Diffusion Unit: Interpretable Edge Enhancement and Suppression Learning for 3D\\Point Cloud Understanding\\
Enhancing Local Feature Learning Using Diffusion for 3D Point Cloud Understanding
} % Replace with your title

% INITIAL SUBMISSION 
% \begin{comment}
\titlerunning{title} 
\authorrunning{Haoyi Xiu 
\and Xin Liu 
\and Weimin Wang
\and Kyoung-Sook Kim 
\and Takayuki Shinohara
\and Qiong Chang
\and Masashi Matsuoka} 

\author{Haoyi Xiu\inst{1,2} 
\and Xin Liu \inst{1}
\and Weimin Wang \inst{2,3}
\and Kyoung-Sook Kim  \inst{2}
\and Takayuki Shinohara \inst{4}
\and Qiong Chang \inst{5}
\and Masashi Matsuoka\inst{1}} 
\institute{Department of Architecture and Building Engineering, Tokyo Institute of Technology, Tokyo, Japan
\and Artificial Intelligence Research Center, AIST, Tokyo, Japan
\and DUT-RU International School of Information Science and Engineering, Dalian University of Technology, Dalian, China
\and Innovation Technology Office Research Center, PASCO Corporation, Tokyo, Japan
\and Department of Computer Science, Tokyo Institute of Technology, Tokyo, Japan}

% \end{comment}
%******************

% CAMERA READY SUBMISSION
\begin{comment}
\titlerunning{Abbreviated paper title}
% If the paper title is too long for the running head, you can set
% an abbreviated paper title here
%
\author{First Author\inst{1}\orcidID{0000-1111-2222-3333} \and
Second Author\inst{2,3}\orcidID{1111-2222-3333-4444} \and
Third Author\inst{3}\orcidID{2222--3333-4444-5555}}
%
\authorrunning{F. Author et al.}
% First names are abbreviated in the running head.
% If there are more than two authors, 'et al.' is used.
%
\institute{Princeton University, Princeton NJ 08544, USA \and
Springer Heidelberg, Tiergartenstr. 17, 69121 Heidelberg, Germany
\email{lncs@springer.com}\\
\url{http://www.springer.com/gp/computer-science/lncs} \and
ABC Institute, Rupert-Karls-University Heidelberg, Heidelberg, Germany\\
\email{\{abc,lncs\}@uni-heidelberg.de}}
\end{comment}
%******************
\maketitle

\begin{abstract}
% The abstract should summarize the contents of the paper. LNCS guidelines
% indicate it should be at least 70 and at most 150 words. It should be set in 9-point
% font size and should be inset 1.0~cm from the right and left margins.

Learning point clouds is challenging due to the lack of connectivity information, i.e., edges. Although existing edge-aware methods can improve the performance by modeling edges, how edges contribute to the improvement is unclear. In this study, we propose a method that automatically learns to enhance/suppress edges while keeping the its working mechanism clear. First, we theoretically figure out how edge enhancement/suppression works. Second, we experimentally verify the edge enhancement/suppression behavior. Third, we empirically show that this behavior improves the performance. In general, we observe that the proposed method achieves competitive performance in point cloud classification and segmentation tasks.

\end{abstract}

\section{Introduction}
% bg 
A 3D point cloud is the most basic shape representation in which the scanned surface is represented as a set of points in the 3D space. With the advent of cost-effective sensors, an increasing number of large-scale point cloud datasets have been released to researchers, facilitating deep learning--based point cloud understanding. Typical applications of such research include autonomous driving~\cite{cui2021deep,qi2018frustum} and remote sensing~\cite{zhu2017deep,shinohara2020fwnet}. 

% regular CNNs -> point CNNs
A point cloud is naturally unordered and unstructured, hindering the application of convolutional neural networks (CNNs) that are suited for processing regular grid data. Therefore, many prior studies have focused on applying CNNs by projecting point clouds to regular grids \cite{kanezaki2018rotationnet,su2015multi,maturana2015voxnet,zhou2018voxelnet}. However, there is information loss incurred by the projection and these approaches are suboptimal. To remedy this issue, PointNet applies shared multi-layer perceptrons (MLPs) and symmetric functions to raw 3D points, consuming point cloud in a lossless manner ~\cite{qi2017pointnet}. PointNet++ subsequently applies PointNets to local subsets of points, obtaining CNN-like translation invariance ~\cite{qi2017pointnet++}. Recently, various convolution methods that operate directly on raw point clouds have been developed~\cite{li2018pointcnn,wu2019pointconv,thomas2019kpconv,liu2019relation}. Despite these efforts, learning point cloud data remains challenging due to the difficulty in inferring the underlying continuous surface from discrete point samples.
%without default connectivity information, i.e., edges. Therefore, much effort is invested into incorporating edge awareness into the network design. 

% previous work--problem pairs 
In recent years, people find that exploring the connectivity information between points (i.e., edges) is beneficial for 3D point cloud understanding and developed edge-aware approaches. 
For instance, researchers treat edges as additional contextual features~\cite{wang2019dynamic,liu2020closer,xiang2021walk} or spatial weights~\cite{wang2019graph,zhao2019pointweb,zhao2021pointtransformer} that describe local geometrical structures and incorporate them into their models. Although incorporating edge information successfully improves the model performance, the underlying mechanism of \textit{how} edges contribute to the improvement is not clear. Moreover, some researchers explicitly supervise the model with edge information~\cite{jiang2019hierarchical,hu2020jsenet,yu2018ec}. However, these methods require per-point and clean ground truth labels (possibly with additional annotations), which are costly and not always available in practice.

In this study, we go beyond edge awareness, and propose the diffusion unit (DU) that performs automatic edge enhancement/suppression learning without additional supervision. Built on the nonlinear diffusion theory~\cite{perona1990scale,weickert1998anisotropic}, DU adaptively enhances task-beneficial edges and suppress irrelevant ones so as to improve the performance. In contrast to existing works, the mechanism of edge enhancement/suppression is interpretable in terms of theoretical analysis and experimental verification. Specifically, 1) We theoretically figure out which component of our method is responsible for edge enhancement/suppression and how it works; 2) We experimentally observe and verify the edge enhancement/suppression behavior; 3) We empirically demonstrate that this behavior contributes to the performance improvement.

%We perform theoretical analysis on DU to investigate its fundamental mechanism regarding edge enhancement/suppression. Furthermore, the close relationship of the diffusion and smoothing~\cite{barash2004common} offers us a systematic way to interpret the effect of DU using smoothness, in which our theoretical analysis is also verified.
% \xiu{KPConv Light }
DU is generally applicable, as it can be seamlessly integrated with a convolution operator as a basic building block of deep neural networks. Particularly, we resort to KPConv~\cite{thomas2019kpconv} as the default convolution operator owing to its superiority in point cloud processing. Further, to better fitting to DU, we develop a lightweight variant of KPConv, namely KPConv-l, which provides decent performance while drastically reducing the number of parameters.

Stacking KPConv-l and DU as a basic building block, we construct DU-Nets to tackle point cloud understanding tasks. Extensive experiments across several standard benchmarks demonstrate its effectiveness. In particular, we achieve the state-of-the-art performance in point cloud classification and comparative performance in part segmentation and scene segmentation.

%To seamlessly integrate DUs into modern deep learning networks, we attach a DU to each convolution layer so that the edge hierarchy is well-exploited. We resort to KPConv~\cite{thomas2019kpconv} as our convolution operator for its simplicity; however, to efficiently tackle with voluminous point clouds, we develop a lightweight version of KPConv, namely KPConv Light (KPConv-l), which improves the performance while drastically reducing the number of parameters.
% \xiu{Network, experiments}
%We construct two networks, coined DU-Nets, by stacking DUs and KPConv-l to tackle point cloud classification and segmentation. Extensive experiments across challenging benchmarks demonstrate the competitive performance and general applicability of DU-Nets. In particular, we achieve state-of-the-art performance in point cloud classification.

Our main contributions are summarized as follows:
\begin{itemize}
    \item We propose DU that performs automatic edge enhancement and suppression learning so as to improve the performance.
    \item We theoretically analyze and experimentally verify the edge enhancement and suppression behavior of DU.  
    \item We propose a lightweight version of KPConv, KPConv-l, which drastically reduces the number of parameters without compromising the performance.
    \item We design DU-Nets by stacking KPConv-l and DU as the basic building block and achieve the state-of-the-art performance in point cloud classification and comparative performance in part segmentation and scene segmentation tasks.
    %KPConv-l, a lightweight convolution operator for point cloud data. It improves the performance while drastically reducing the number of parameters.
    %\item Extensive experiments verify the generalization capability of DU-Nets. In particular, DU-Net achieves the state-of-the-art performance in point cloud classification.  
\end{itemize}

\begin{figure}[t]
    \centering 
        \includegraphics[width=0.9\linewidth]{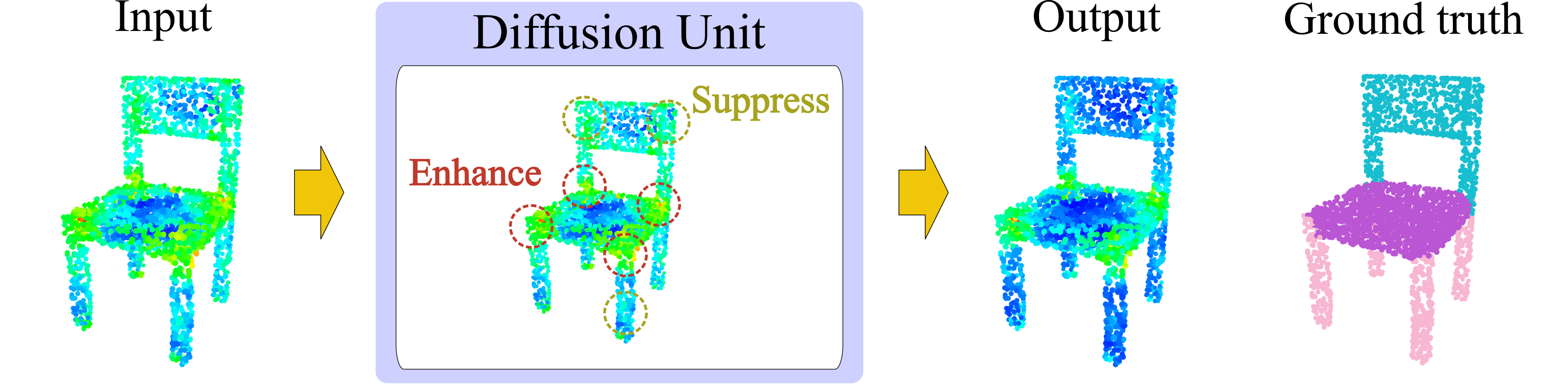}
    \caption{Overview of the diffusion unit (DU). DU automatically learns to enhance task-beneficial edges or suppress irrelevant edges so as to improve the performance.}
    \label{fig: overview}
\end{figure}
\section{Related Work}
% In this section, we provide a brief review of related fields \wm{have a check whether the added sentences here are correct. that aim to understand point clouds/images from the structure-ware and efficiency aspects}, including deep learning approaches on 3D point clouds understanding, structure-aware methods and efficient convolutions.
\subsection{Deep learning for 3D Point Clouds}
\textbf{Projection-based methods} Projection-based methods project point clouds to regular grids (e.g., 2D planes~\cite{kanezaki2018rotationnet,su2015multi,feng2018gvcnn} and 3D voxels~\cite{maturana2015voxnet,zhou2018voxelnet,graham20183d,choy20194d}) to make matured regular convolution applicable to point clouds. However, they lose fine-grained details through projections. 

\noindent\textbf{MLP-based methods} Pioneered by PointNet~\cite{qi2017pointnet}, MLP-based methods prevent information loss by operating directly on the raw points. PointNet relies on shared MLPs and symmetric functions; both operations are permutation-invariant, and thus the irregularity of point clouds is well-resolved. Subsequently, PointNet++~\cite{qi2017pointnet++} have made major steps toward the convolution-like operation by applying shared MLPs to local subsets of points~\cite{zhang2019shellnet,lan2019modeling}. 

\noindent\textbf{Convolution-based methods} A variety of point convolutions have been realized by defining convolution operations on unstructured point clouds. Some studies construct regular kernels/grids on which the points are projected, thus enabling the faithful extension of the standard convolution~\cite{thomas2019kpconv,mao2019interpolated}; the others dynamically generate convolution filters based on positional features~\cite{li2018pointcnn,wu2019pointconv}. Though being effective, convolution-based methods do not explicitly model local structures which potentially can provide more accurate representations of the underlying surface.  
% \vspace{-5pt}

\noindent\textbf{Edge-aware methods} Edge-aware methods explicitly integrate edge information into the network design. 
% In general, there are three types of edge-aware approaches. 
Pioneered by EdgeConv~\cite{wang2019dynamic}, these methods perform convolution on the edge embedding to better model the local geometric structure. The idea of EdgeConv is adopted in numerous subsequent works (e.g., \cite{simonovsky2017dynamic,li2019deepgcns,liu2020closer,xu2021paconv,xiang2021walk}). Although performance improves, it is difficult to analyze rigorously how it is improved. Furthermore, local edge features are simply fused with global ones by MLPs, which may be suboptimal since they are significantly distinct features.
On the other hand, some methods regard edge information as a similarity measure, representing the semantic distance. In ~\cite{wang2019graph}, such a similarity is used for describing connectivity among neighboring points; consequently, the information exchange is guided by edge information. Such an operation is often coupled with the attention mechanism~\cite{zhao2019pointweb,zhao2021pointtransformer,xiu2022enhancing}, which converts edge into normalized spatial weights. However, the forced conversion may lose rich structural information (e.g., smoothness) contained in the edges.
Another type of approach explicitly guides the network by edge-related supervision. Specifically, networks are taught to maintain spatial consistency~\cite{jiang2019hierarchical}, perform edge detection and other tasks jointly~\cite{hu2020jsenet} or be aware of the location of edges~\cite{yu2018ec}. Such methods involve dedicated loss functions or models, and often require clean and per-point ground truth, thereby making their practical applications challenging.

In contrast to the aforementioned edge-aware methods, the proposed DU goes beyond edge awareness by automatically learning to enhance task-beneficial edges and suppress irrelevant ones, thereby improving the performance. 
Furthermore, compared with existing edge-aware methods, DU offers significantly better interpretability through both theoretical and qualitative analysis.

\subsection{Diffusion}
In essence, diffusion methods, which are motivated by the diffusion equation, model the smoothing of data (e.g., images) as diffusion processes. The core of the diffusion methods is the diffusivity~\cite{perona1990scale,black1998robust}, which is often defined as a function of edge. As a result, diffusion methods remove small edges (small edges) while preserving significant edges~\cite{perona1990scale}, making them attractive to various applications. Therefore, such techniques are extensively studied in image processing~\cite{perona1990scale,weickert1998anisotropic,weickert1999coherence,brox2006nonlinear} and later by other communities (e.g., computer graphics~\cite{desbrun1999implicit,clarenz2000anisotropic,bajaj2003anisotropic}). 
% deep learning--based 
In the context of deep learning, although several works model the global information propagation~\cite{atwood2016diffusion,zhao2021adaptive,chamberlain2021grand} or the probabilistic point cloud generation~\cite{luo2021diffusion} as diffusion processes, extending the diffusion equation for adaptive edge enhancement/suppression for 3D point cloud understanding, which we exclusively cope with in this study, remain unexplored.

\section{Diffusion Unit}
In this part, we first provide a brief background about the diffusion equation to build intuition. 
Next, we present the (continuous) definition of DU.
Then, we perform a theoretical analysis to reveal the underlying mechanism of edge enhancement/suppression learning.
Subsequently, we provide the discretization scheme that enables an efficient implementation of DU on modern machines. To enable a seamless integration of DUs to modern CNN-based frameworks, we explain how DUs can be integrated with KPConv-l. Lastly, we describe the network architectures used for various analyses and experiments in this study.   
\subsection{Preliminary}
The diffusion equation describes the movement of diffusive substances from regions of higher concentration to lower concentration without creating or destroying mass~\cite{weickert1998anisotropic}. For instance, when hot water is poured into cold water, heat diffuses until the temperature of the water becomes the same everywhere. Let ${u}({p}, t)$ denote the concentration at the position ${p}$ and time $t$. 
% The diffusion process is describe by the following partial differential equation:
The amount of substances that flow through per unit area per unit time (the flux) is described by Fick's law:
\begin{equation}
    {s} = - g \cdot \nabla {u} \;, 
    \label{eq: flux}
\end{equation}
where $\nabla$ denotes the gradient operator and $g$ denotes the diffusivity. The fact that diffusion processes do not create or destroy mass is expressed by the continuity equation:
\begin{equation}
    \partial_t {u} = - \mathrm{div}({s}) \;.
    \label{eq: continuity}
\end{equation}
The continuity equation indicates that the change of concentration is caused only by the flux, which is measured by the divergence operator ($\mathrm{div}$). Finally, the diffusion process is described by combining the above two equations: 
\begin{equation}
    \partial_t {u} = \mathrm{div} (g \cdot \nabla {u}) \;\;\; t \ge 0\;,
    \label{eq: diffusion_equation}
\end{equation}
with the initial condition ${u}({p},0)={u}_0({p})$ and the boundary condition as appropriate.

% \subsubsection{Definition of the Diffusion Unit}
% \vspace{3pt}
% \noindent\textbf{Definition of the Diffusion Unit}
\subsection{Definition of Diffusion Unit}
Inspired by the diffusion equation, we propose diffusion unit (DU) that facilitates the edge enhancement learning for 3D point clouds. Suppose a continuous spatial-temporal multi-channel point cloud $\mathbf{u} = \mathbf{u}(\mathbf{p}, t) = (u_1(\mathbf{p}, t), u_2(\mathbf{p}, t), ..., u_d(\mathbf{p}, t))$, where $d$ is the number of channels, $t$ is time, $\mathbf{p}$ denotes the position vector, and the initial condition is $\mathbf{u}(\mathbf{p}, 0) = \mathbf{h}$.
%The input is a multi-channel point cloud $\mathbf{h}=(h_1(\mathbf{p}), h_2(\mathbf{p}), ..., h_d(\mathbf{p}))$, where $d$ is the number of channels and $\mathbf{p}$ denotes the position vector. 
%$\mathbf{u}(\mathbf{p}, t)$ denote the feature of the point cloud at the position $\mathbf{p}$ and time $t$.
DU is defined as:
% \begin{equation}
%         \partial_t u_i = \mathrm{div}(\phi_i(\nabla u)), \;\;\; t\ge0,
%     \label{eq: continuous_DU}
% \end{equation} 
% $\nabla u$ encodes channel-wise spatial gradient, $i=1,2,..., d$ indexes the output channel, and $\phi_i:\mathbb{R}^d \rightarrow \mathbb{R}$ represents a neural network. As opposed to previous works that adopt a common diffusivity to all channels~\cite{gerig1992nonlinear,chamberlain2021grand} to control the local smoothing behavior, Eq.~\eqref{eq: continuous_DU} states that $i$th output channel is dependent on \textit{all} input channels of $\nabla u$. Assuming that each input channel contains the distinct structural information of different importance, we believe that a dedicated filter ($\phi_i$) would find a optimal combination of channels to detect and enhance important structures.

\begin{equation}
        \partial_t \mathbf{u} = \mathrm{div}(\phi(\nabla \mathbf{u})), \;\;\; t\ge0 \;,
    \label{eq: continuous_DU}
\end{equation} 
where $\partial_t \mathbf{u} \in \mathbb{R}^d$ is the output of DU at time $t$, $\nabla \mathbf{u}$ encodes channel-wise spatial gradient.
%and $\phi:\mathbb{R}^d \rightarrow \mathbb{R}^d$ represents a fully-connected neural network.
Note that the choice of diffusivity $g$ in Eq.~\eqref{eq: diffusion_equation} has a significant impact on performance. Finding the appropriate $g$ often requires domain knowledge and involves numerous trials and errors ~\cite{black1998robust}. In our definition, we replace the handcrafted $g$ with a \textit{trainable filter} $\phi:\mathbb{R}^d \rightarrow \mathbb{R}^d$ so that the diffusivity function can be learned w.r.t data and task.
% In most previous works that deal with multi-channel inputs~\cite{gerig1992nonlinear,chamberlain2021grand}, a common diffusivity is applied to all channels, i.e., no channel mixing, so that their behaviors are synchronized during diffusion processes \liu{Do we really need to mention this? Mixing channel is indeed not a big contribution, but if we write in this way, I have a feeling that similar approaches have been proposed in other fields. I think we can just state that assuming that each channel encodes a fragment of structural information with various degrees of noise, we transform ...}; 
In practice, we use \texttt{1D-Conv} to implement $\phi$ to maintain the local behavior. %so that weights are shared across all input elements. 
Note that $\phi$ is a multi-channel filter that mixes information in all channels, since each channel may contain a fragment of information with certain degrees of noise. %\xiu{may be this: Note that $\phi$ mixes information in all channels to produce refined ones, since each channel may contain a fragment of information that encodes various aspects of local structures. A weighted combination of fragments propagates information to all channels such that the resulting representation is appropriately calibrated.}
%Assuming that each channel encodes a fragment of structural information with different degrees of noise, we transform the input with a trainable filter ($\phi$) that mixes all channels to produce refined ones. 
% As opposed to previous works that adopt a common diffusivity to all channels~\cite{gerig1992nonlinear,chamberlain2021grand} to control the local smoothing behavior, Eq.~\eqref{eq: continuous_DU} states that $i$th output channel is dependent on all input channels of $\nabla u$. Assuming that each input channel contains the distinct structural information of different importance, we believe that a dedicated filter ($\phi_i$) would find a optimal combination of channels to detect and enhance important structures.

% \subsubsection{Structure Awareness}
% \vspace{3pt}
% \noindent\textbf{Edge Awareness}
\subsection{Edge Enhancement/Suppression Learning}
\label{sec: theoretical_analysis}
% \begin{floatingfigure}[r]{0.25\textwidth}
%     \includegraphics[width=0.25\textwidth]{ECCV/figures/profiles_vertical.png}
%     \caption{Caption}
%     \label{fig:my_label}
% \end{floatingfigure}
% \begin{wrapfigure}{r}{0.3\linewidth}
%     \centering 
%         \includegraphics[width=1.0\linewidth]{ECCV/figures/diffs.png}
%         \vspace{-20pt}
%     \caption{edge profile}
%     \label{fig: edge_profile}
% \end{wrapfigure}
Now we explain how DU performs edge enhancement/suppression learning.
%In contrast to classic edge detection methods that detects all edges, DU strives to detect edges in a need-dependent manner, that is, it enhances salient edges that are beneficial for minimizing the training objective while suppressing non-essential ones, using a neural network.

\begin{wrapfigure}{r}{0.3\textwidth}
    % \vspace{-20pt}
    \centering
    \includegraphics[width=0.3\textwidth]{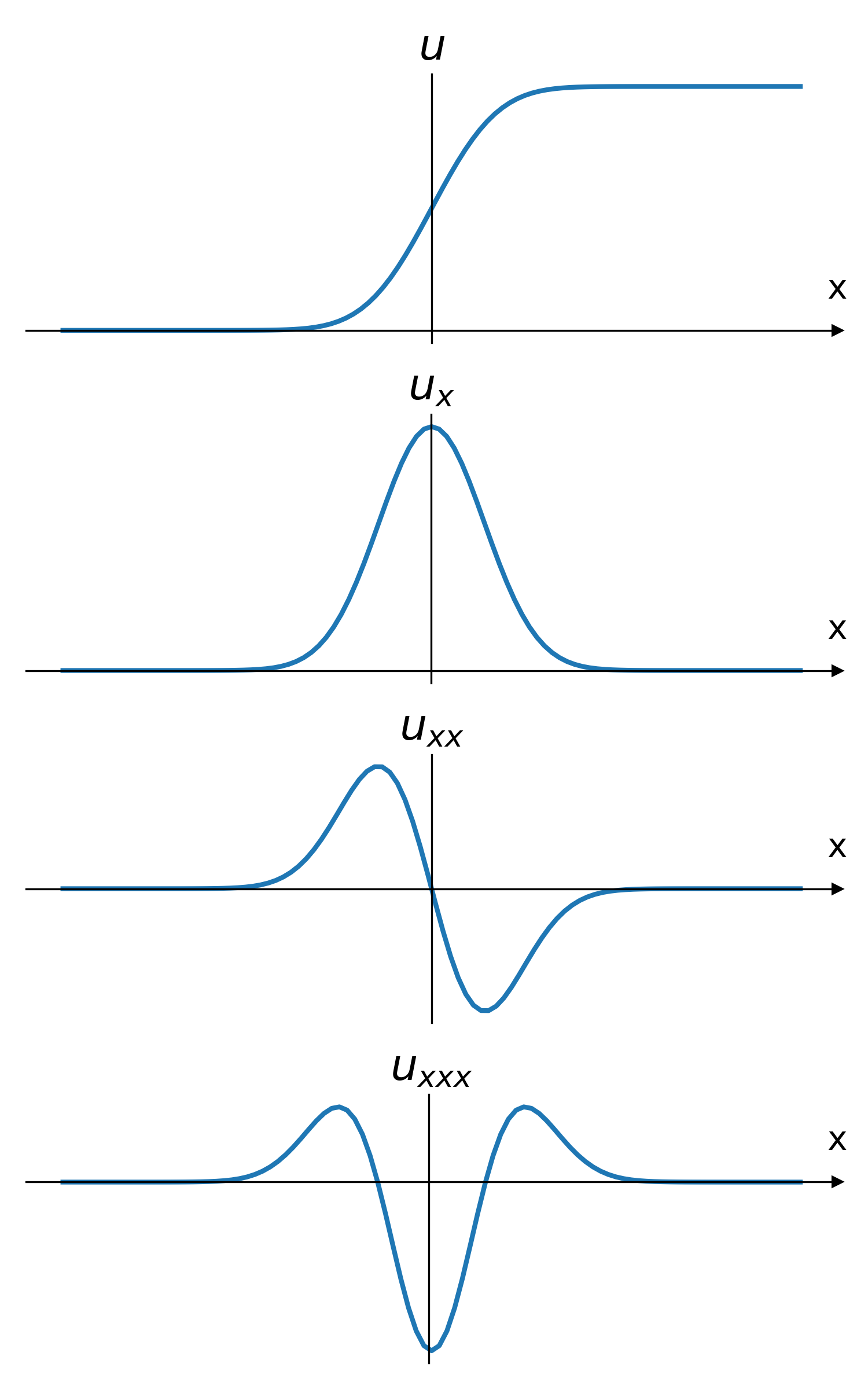}
    \caption{Profiles of the smoothed step edge and its derivatives.}
    \label{fig: edge_profile}
    \vspace{-80pt}
\end{wrapfigure}

For simplicity, we consider a step edge convolved by a Gaussian. Ideally, such a structure can be detected by the spatial gradient $\nabla \mathbf{u}$. Without loss of generality, we assume that the edge is aligned with $x$ axis ($\mathbf{u}_y=\textbf{u}_z=\textbf{0}$). The profile of the step edge and its derivatives are illustrated in Fig.~\ref{fig: edge_profile}. Now we focus on a single output channel $i$ for brevity.

% Among various types of structures, we focus on analyzing step edge convolved by a Gaussian for simplicity. Ideally, such a structure can be reliably detected by analyzing the concentration gradient $\nabla u$. The profile of the edge and its derivatives are shown in Fig.~\ref{fig: edge_profile}.

% \begin{figure}
%     \centering
%     \includegraphics[width=0.9\linewidth]{ECCV/figures/profiles.png}
%     \caption{Profiles of the smoothed step edge and its derivatives.}
%     \label{fig: edge_profile}
% \end{figure}

% We aim to detect the structure that is characterized by the gradient. 
% For instance, the region boundary on which feature values change rapidly is a detectable structure. 
% To investigate the behavior of DU near such a structure, we model an ideal structure using the error function as shown in Fig.~\ref{fig: diffs}.
% Furthermore, we say the edge is enhanced when the magnitude of the gradient is increased and smoothed when the magnitude of the edge is decreased.

In this context, Eq.~\eqref{eq: continuous_DU} can be simplified as
\begin{align}
    (\partial_t \textbf{u})_i
        &=
        \frac{\partial}{\partial x}\left(
            \phi_i(\textbf{u}_x)
            \right)
            =
            \nabla\phi_i \cdot \textbf{u}_{xx}\\
        &= \sum_{j=1}^{d} (\phi_i')_j \cdot \left(\textbf{u}_{xx}\right)_j, \;\;\; i=1,2,...,d \;.
\end{align}

We are interested in the evolution of the edge over time, i.e., how $\textbf{u}_x$ changes during applications of DU. We can derive that:

\begin{align}
        (\partial_t \textbf{u}_x)_i
        &= \left(
                \frac{\partial}{\partial t}\frac{\partial \textbf{u}}{\partial x}
            \right)_i
        =\frac{\partial}{\partial x}
                    \left(\frac{\partial \textbf{u}}{\partial t}\right)_i
                    \\
        &= \frac{\partial}{\partial x}\left(
            \sum_{j=1}^{d}(\phi_i')_j
            \cdot (\textbf{u}_{xx})_j
        \right) \\
        &= \sum_{j=1}^{d} (\phi_i^{''})_j \cdot (\textbf{u}_{xx})_j^2 + (\phi_i')_j \cdot (\textbf{u}_{xxx})_j \; \label{eq: contribution}.
\end{align}

As shown in Fig.~\ref{fig: edge_profile}, at the inflection point $\textbf{u}_{xx} = 0$ and $\textbf{u}_{xxx} < 0$. Therefore, the contribution of a particular input channel $j$ to the output channel $i$ (the sign of Eq.~\eqref{eq: contribution}) is determined by the sign of $(\phi_i')_j$. Specifically, channel $j$ has a positive impact if $(\phi_i')_j < 0$, whereas it has a negative impact if $(\phi_i')_j > 0$.

As a result, enhancing ($(\partial_t\textbf{u}_x)_i > 0$) or suppressing ($(\partial_t\textbf{u}_x)_i < 0$) the edge can be adaptively learned by the filter $\phi$, by collectively using fragmentary information in all channels.
%collectively determined by all input channels. In such a way, we expect the trainable filter $\phi$ to refine each channel adaptively using fragmentary information in all channels.
% Consequently, we expect that the significant structure would be detected and enhanced by summing up contributions from each input channel. 

\subsection{Discretization}
\label{sec:discretization}
\begin{figure}[t]
    \centering
    \includegraphics[width=0.9\linewidth]{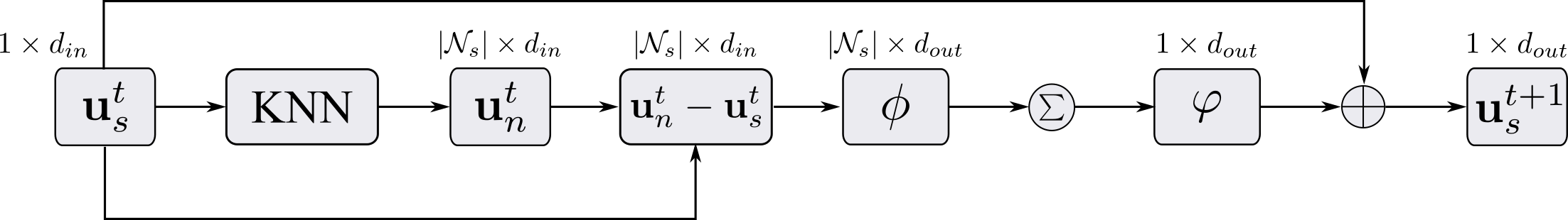}
    \caption{
    The computation flow of DU. The input dimension is the same as the output dimension ($d_{in}=d_{out}$). $\bigoplus$ indicates the element-wise addition.
    % The computation flow of DU (left) and LightPConv (right). The input channel dimension is the same as output one ($d_{in}=d_{out}$). $\bigoplus$ represents the elementwise addition. $\bigotimes$ represents the matrix multiplication.
    }
    \label{fig: computation_flow}
\end{figure}

% \vspace{3pt}
% \noindent\textbf{Discretization}
To deal with discrete point clouds, discretization of Eq.~\eqref{eq: continuous_DU} is necessary. Let $\textbf{u}_s, \textbf{u}_n \in \mathbb{R}^d$ denote the feature of the center point and its spatial neighbors, respectively. Let $n \in \mathcal{N}_s$ represents $n$-th neighbor of the center point. First, we adopt the explicit scheme for time discretization and have:
\begin{equation}
    \partial_t \textbf{u}_s    
    \approx \textbf{u}_s^{t+1} - \textbf{u}_s^t \;. 
    \label{eq: time_discretization}
\end{equation}
Second, we discretize the space using the finite difference and have:
\begin{equation}
    \mathrm{div}(\phi(\nabla \textbf{u}))
    \approx \frac{1}{|\mathcal{N}_s|} \sum_{n \in \mathcal{N}_s} \phi\left(\textbf{u}_n^t - \textbf{u}_s^t \right)\;,
    \label{eq: spatial_discretization}
\end{equation}
where $|\mathcal{N}_s|$ represents the number of neighbors. %$\frac{1}{|\mathcal{N}_s|}\sum_{n\in\mathcal{N}_s\phi(\cdot)}$ and $\textbf{u}_n^t - \textbf{u}_s^t$ are the discrete analogies of the gradient $\nabla \textbf{u}$ and divergence $\textrm{div}(\cdot)$.  
Combining Eqs.~\eqref{eq: continuous_DU}, \eqref{eq: time_discretization}, \eqref{eq: spatial_discretization}, we have
\begin{equation}
    \textbf{u}_s^{t+1}
    =  \textbf{u}_s^t + 
    \left(\frac{1}{|\mathcal{N}_s|} \sum_{n \in \mathcal{N}_s} \phi\left(\textbf{u}_n^t - \textbf{u}_s^t\right)\right)\;.
        \label{eq: DU_discretization_pre}
    \end{equation}
Moreover, we incorporate Batch Normalization and ReLU activation function $\varphi = \texttt{BatchNorm}\cdot\texttt{ReLU}: \mathbb{R}^d \rightarrow \mathbb{R}^d$ into the second term on the right-hand side of Eq.~\eqref{eq: DU_discretization_pre} to facilitate training and encourage sparsity. 
%incorporate Eq.~\eqref{eq: spatial_discretization} into $\varphi:\mathbb{R}^d \rightarrow \mathbb{R}^d$, in which the input is transformed by a Batch Normalization~\cite{ioffe2015batch} and an activation function (e.g., ReLU), to stabilize and facilitate the training while encouraging the sparsity. 
% here add the consequence of adding relu ? 
Finally, the discretized DU is defined as:
\begin{equation}
    \textbf{u}_s^{t+1} 
    = \textbf{u}_s^t + 
        \varphi \left(\frac{1}{|\mathcal{N}_s|} \sum_{n \in \mathcal{N}_s} \phi\left(\textbf{u}_n^t - \textbf{u}_s^t\right)\right)\;.
\end{equation}
Note that the neural network functions $\phi$ and $\varphi$ work together and are responsible for the learning of enhancing or suppressing edges (we provide qualitative analysis on the behavior of $\phi$ and $\varphi$ in Sec.~\ref{sec: interpretation}). The above definition can be efficiently computed and easily parallelized, fitting to modern GPU-empowered deep learning frameworks. 
%\liu{remove this sentence: Notably, the discretized DU bears some similarity to renowned ResNets~\cite{he2016deep} by possessing an identity connection, which would be beneficial for backpropagation.} 
The computation flow of DU is shown in Fig.~\ref{fig: computation_flow}. 

\subsection{Integrating with a Lightweight Point Convolution}
DU can be integrated with a convolution operator to build a basic building block of a deep neural network. In particular, we choose KPConv~\cite{thomas2019kpconv} as the default convolution operator, owing to its superiority in point cloud processing. However, it suffers from high memory consumption, hindering its application to voluminous data. Therefore, we further develop a lightweight version of KPConv (KPConv-l) to achieve decent performance while drastically reducing the model parameters.

\begin{figure}[t]
    \centering
    \includegraphics[width=0.9\textwidth]{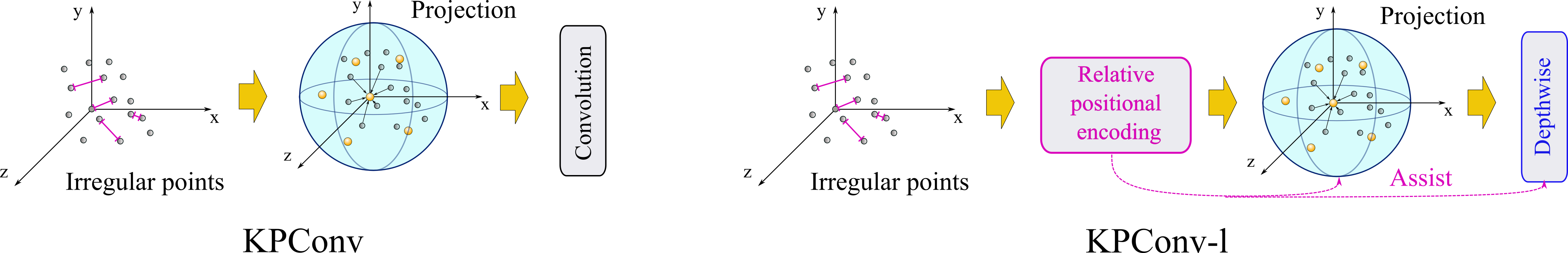}
    \caption{Differences between KPConv~\cite{thomas2019kpconv} and KPConv-l. KPConv-l reduces the complexity by replacing standard convolution with the depthwise one. Further, to assist the capacity-limited depthwise convolution to handle irregular spacing of points, relative positional encoding is applied to the raw input points.}
    \label{fig:difference_KPConv-l_KPConv}
\end{figure}

%so that it achieves higher accuracy while drastically reducing parameters.
%A seamless integration of DUs into modern deep learning network is necessary to achieve competitive performance. 
%In particular, CNN is a strong architecture when dealing with vision related tasks. In particular, we integrate DU with a convolution operator to build the basic building block of the network.
%We choose KPConv rigid (KPConv-r)~\cite{thomas2019kpconv} as our convolution operator; more importantly, we customize KPConv so that it achieves higher accuracy while drastically reducing parameters.

\vspace{3pt}
\noindent\textbf{Lightweight KPConv} 
% describing KPConv
KPConv adapts the standard convolution for regular data to the point cloud setting by constructing artificial convolution kernels, to which the input points are projected. However, the resulting convolution is parameter-consuming, which limits its applications to voluminous point clouds. Specifically, let $l$ denote the number of neighbors involved in the convolution. Further, let $d_{in}, d_{out}$ denote the dimensions of input channel and output channel, respectively. The standard convolution requires $l\times d_{in} \times d_{out}$ parameters, which leads to excessive memory consumption. To remedy this issue, we borrow the idea from prevalent depthwise separable convolution (DSC)~\cite{sifre2014rigid} to simplify the KPConv into depthwise KPConv (with a depth multiplier~\cite{howard2017mobilenets}). As a result, the number of parameters is reduced to $l\times d_{out}$, thereby significantly reducing the memory consumption.

A potential concern is that, unlike data that have regular spacing between elements (e.g., image), point clouds typically have irregular spacing, which poses a great challenge to capacity-limited depthwise KPConv. Therefore, we apply relative positional encoding on the \textit{raw} input points to assist depthwise KPConv to learn the irregular spacing. Specifically, relative positional encoding transforms the position-concatenated point features by an MLP such that subsequent operations become position-aware. 
% adopt a distinct strategy, and apply pointwise convolutions \textit{first} to the raw input features of neighboring points (before projection) to assist the learning of irregular spacing. Subsequently, the resulting input features are projected to convolution kernels and processed by depthwise KPConv. 
Though simple, this trick effectively reduces memory consumption without compromising the performance. The differences of KPConv and KPConv-l are shown in Fig.~\ref{fig:difference_KPConv-l_KPConv}.

\vspace{3pt}
\noindent\textbf{Integration Strategy} In this study, we stack a KPConv-l and a DU to obtain the feature representation. Concretely, the input is transformed by a KPConv-l, which is followed by a DU. Popular architectures in point cloud processing typically down-sample the input in a convolution layer; therefore, attaching DU to each convolution layer facilitates the network to learn the multi-resolution edge hierarchy.

\subsection{Network Architecture}
\begin{figure}
    \centering
    \includegraphics[width=0.85\textwidth]{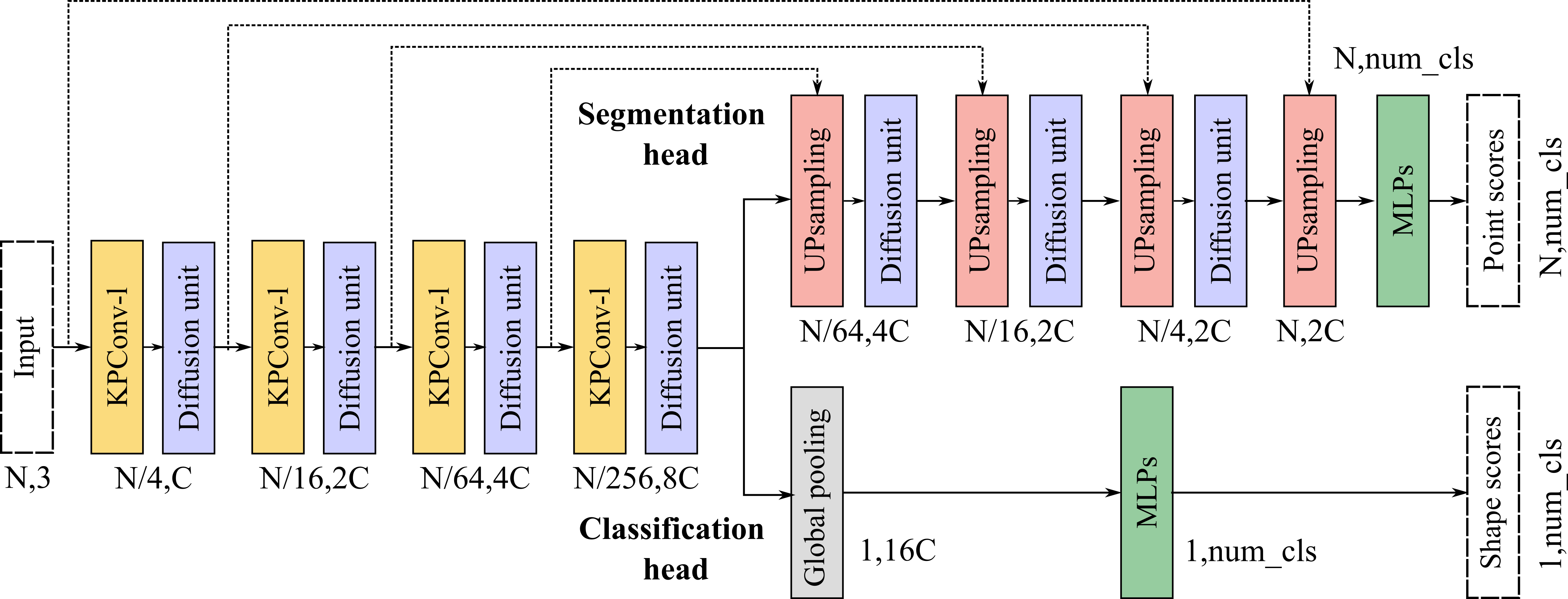}
    \caption{The architecture of the model used in this study. N, C denote the number of input points and channels, respectively. }
    \label{fig: model}
\end{figure}
In this study, we tackle point cloud classification and segmentation. We construct Diffusion Unit--Enhanced Networks (DU-Nets) by stacking KPConv-l and DU as the basic building block. 
For classification, we follow multi-scale PointNet++~\cite{qi2017pointnet++} to construct an encoder composed of four pairs of KPConv-l and DU. Each layer downsamples the input point cloud using the furthest point sampling~\cite{qi2017pointnet++}. The global representation is obtained by applying a pooling layer to the last layer of the encoder, and is subsequently fed to the MLPs to generate the class scores. For segmentation, we adopt the same encoder as the classification model. Then, the output of the encoder is successively upsampled to the original resolution using 3-nearest neighbor upsampling layers. Notably, each interpolation layer is followed by a DU such that salient edges are kept sharpened. The final per-point scores are obtained by feeding the output of the decoder to a series of MLPs. In addition, U-Net~\cite{ronneberger2015u}-like skip connections are used to assist the feature upsampling. The architecture is illustrated in Fig.~\ref{fig: model}.

% \vspace{3pt}
% \noindent\textbf{Encoder.}
% Following PointNet++~\cite{qi2017pointnet++}, we construct the encoder as a stack of DALCs. Each DALC (excluding the last one) is followed by a uniform sampling that effectively abstracts the resolution of input points. As a result, the encoder achieves multi-resolution learning. Furthermore, LightPConvs operate on different scales using different radii in each DALC, such that the encoded features encapsulate multi-scale information.        
%

% \paragraph{Decoder.} 
% \vspace{3pt}
% \noindent\textbf{Decoder.} The decoder is constructed by stacking DU-augmented feature propagation layers~\cite{qi2017pointnet++}. Specifically, coarse features are upsampled using nearest-neighbor interpolation followed by a DU. In this manner, features are successively upsampled to a higher resolution while being structure-aware. 
%

% \paragraph{Models.} 
% \vspace{3pt}
% \noindent\textbf{Models.}
% % The overall model architectures are shown in ~\ref{fig: arch}. 
% We construct two models, DALC-Nets, for the classification and segmentation tasks. For classification, an encoder is followed by a pooling layer to extract a global representation of the input points. The global feature is subsequently fed to the MLPs to generate the class scores. For segmentation, the output of the encoder is successively upsampled to the original resolution using a decoder. U-Net~\cite{ronneberger2015u}-like skip connections are used to assist the feature upsampling. 

\section{Experiment}
\label{sec: exp}
In this section, we conduct experiments to answer the following questions:
\begin{enumerate}[Q1.]
\item Does DU really perform edge enhancement/suppression?
\item How much does DU contribute to improve the performance?
\item Is DU-Net better than existing deep leaning models?
\item Is the design of the different components of DU-Net reasonable?
\end{enumerate}
To answer these questions, we use standard benchmark tests on point cloud classification, part segmentation, and scene segmentation tasks.

%In this section, we first examine the behavior of DU using smoothness. Then, we perform comprehensive experiments to demonstrate the effectiveness of DU-Nets. Specifically, we report the performance of DU-Nets on point cloud classification, part segmentation, and scene segmentation. 

For classification, we use ScanObjectNN~\cite{uy2019revisiting}, which is a challenging dataset consisting of real-world 3D scans. In total, it contains 15k objects, each being labeled into one of the 15 categories.

For part segmentation, we use ShapeNet Part dataset~\cite{yi2016scalable}, which includes 16,880 models 3D models. It includes 16 object classes and 50 object parts, each of which is annotated into two to six parts. For a fair comparison, we use the data provided by \cite{qi2017pointnet++}. 

For scene segmentation, Stanford large-scale 3D indoor spaces (S3DIS)~\cite{armeni20163d} is used to measure the performance. In total, it has 272 indoor environments where each point is assigned a class out of 13 classes.  

% For classification, we use ScanObjectNN~\cite{uy2019revisiting} is used to validate the effectiveness of the DU-Net. The dataset consists of real-world 3D scans; thus, it is more challenging than the synthetic one~\cite{wu20153d}. There are 15k objects included in the dataset in total, and each object is labeled as one of the 15 classes. Each point cloud includes some measurement errors, occlusions, and background points. We use the hardest set of the dataset and adopt the official train-test split, where 80\% of the data are used for training and the remaining 20\% for the test. 
\subsection{Verifying the Behaviors of DU}
\label{sec: interpretation}
\begin{figure}[t]
    \centering 
        \includegraphics[width=0.8\linewidth]{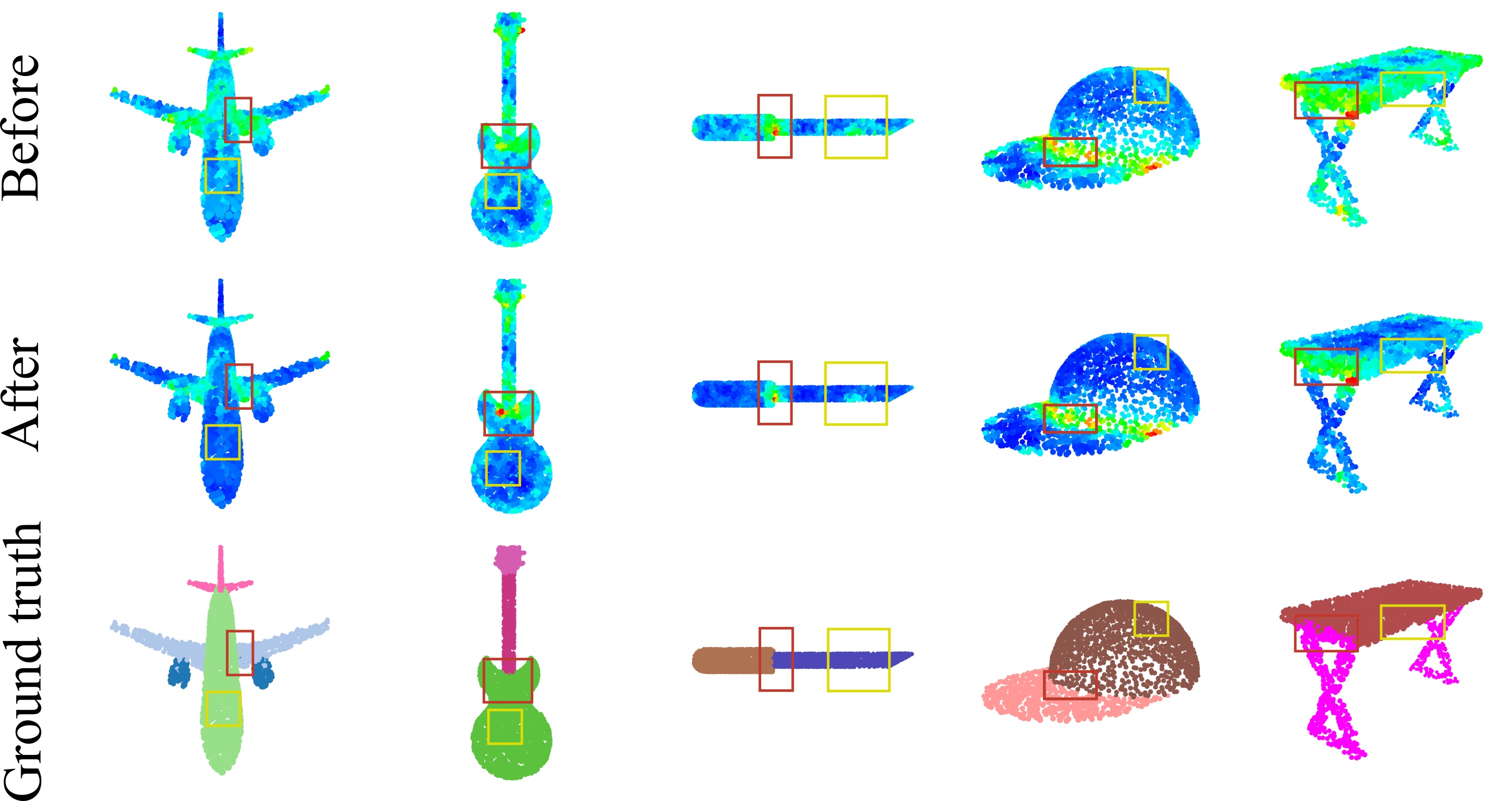}
    \caption{Visualization of the smoothness using ShapeNet dataset (part segmentation).
    % (ShapeNet).
    \textcolor{red}{Red} rectangles show the examples of enhanced part, while \textcolor{Goldenrod}{yellow ones} show the suppressed part. DU successfully enhances the part boundaries while smoothing out other edges.}
    \label{fig: local_smoothness_shapenet}
\end{figure}
\begin{figure}[t]
    \centering 
        \includegraphics[width=0.8\linewidth]{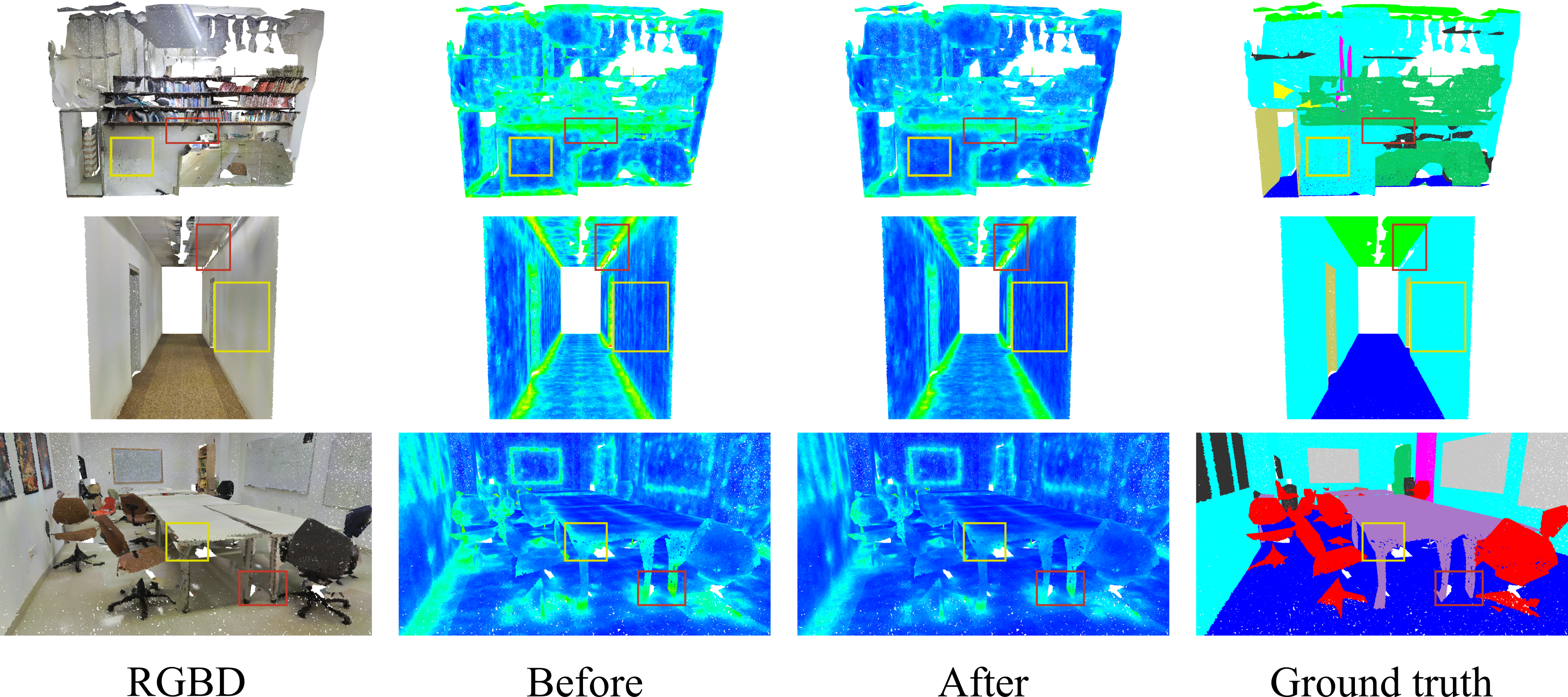}
    \caption{Visualization of the smoothness using S3DIS dataset (scene segmentation). \textcolor{red}{Red} rectangles show the examples of enhanced part, while \textcolor{Goldenrod}{yellow ones} show the suppressed part. DU manages to enhance the task-related edges while suppressing irrelevant ones (e.g., the table in the last row).}
    \label{fig: local_smoothness_s3dis}
\end{figure}
We have theoretically analyzed that the neural network functions $\phi$ and $\varphi$ work together and are responsible for the learning of enhancing or suppressing edges. Here we experimentally verify the edge enhancement/suppression behaviors. Specifically, we perform qualitative analysis on \textit{smoothness}, which reflects the effects of DU.
% diffusion~\cite{barash2004common}. 
Let $\textbf{f}\in\mathbb{R}^d$ denote the processed features. Formally, the smoothness is defined as: $||\sum_{n\in\mathcal{N}_s}\textbf{f}_n - \textbf{f}_s||$, which essentially summarizes how the center point is different from its neighbors. As such, we compute the smoothness before and after applying DU to analyze the behavior of DU.

%We dissect DU into components so that the role of each component can be interpreted. As such, we compute the smoothness of input, output, features after applying $\phi$ and $\varphi$, respectively. We use two models (from Sec.~\ref{sec: part_seg} and \ref{sec: scene_seg}) trained on ShapeNet~\cite{yi2016scalable} and S3DIS~\cite{armeni20163d} visualize the smoothness. 
 
Fig.\ref{fig: local_smoothness_shapenet} and \ref{fig: local_smoothness_s3dis} show several examples from ShapeNet and S3DIS datasets on which our part and scene segmentation models are trained. For each example, smoothness distributions before and after DU are extracted and compared. 
We find that 1) DU successfully enhances the part boundaries and smooths out other intra-region edges in the part segmentation task; 2) DU manages to enhance inter-category edges while suppressing intra-category ones, making object boundaries more salient. Therefore, the task-beneficial edge enhancement/suppression behavior of DU can be verified.

\subsection{Point Cloud Classification}
\label{sec: classification}

% \paragraph{Dataset.}

% ScanObjectNN~\cite{uy2019revisiting} is used to validate the effectiveness of the DU-Net. The dataset consists of real-world 3D scans; thus, it is more challenging than the synthetic one~\cite{wu20153d}. There are 15k objects included in the dataset in total, and each object is labeled as one of the 15 classes. Each point cloud includes some measurement errors, occlusions, and background points.
% We use the hardest set of the dataset and adopt the official train-test split, where 80\% of the data are used for training and the remaining 20\% for the test.
% \vspace{-5pt}

% \paragraph{Configuration.}
We use the most difficult set of the dataset and adopt the official train-test split~\cite{uy2019revisiting}.
The performance is measured by the overall accuracy (OA). We use Adam~\cite{kingma2015adam} optimizer with an initial learning rate of 0.001. The input is augmented by random rotation, scaling, and translation. Only the 3D coordinates are used as input features. Furthermore, we vary the input number of points (1,024 and 2,048) to investigate the impact of the increased training data.
% and trained the model for 250 epochs with a batch size of 32. The initial learning rate is set to 0.001 and decays by a factor of 10 when it plateaus. 
% For a fair comparison, we first evaluate our method using the same protocol as \cite{uy2019revisiting}. 
% The input is augmented by the random rotation, scaling, and translation. Only the 3D coordinates are used as input features. Furthermore, we vary the input number of point (1,024 and 2,048) to investigate the impact of the increased training data.

% \vspace{-5pt}
% \paragraph{Result.} 
% \vspace{3pt}
% \noindent\textbf{Result.}
The result is listed in Table~\ref{tab: result_cls}. The DU-Nets outperform the previous leading methods by significant margins under both experimental settings, which verifies their effectiveness on the classification task. We observe that increasing the number of input points significantly improves the performance. We conjecture that increased density provides a better approximation of the underlying surface, thus leading to a significant improvement. 
\begin{table}[t]
    \small
    \centering
    \caption{Results of point cloud classification on ScanObjectNN dataset. The best, average, and standard deviations of our results in three runs are reported
    % Ours* represents the performance of DC-CNN under the same training protocol as \cite{uy2019revisiting}.
    }
    \begin{tabular}{lccl}
        \hline
        Method
        & \#point
        & 
        & OA
        \\
        \hline
        PointNet~\cite{qi2017pointnet}
        & 1,024
        &
        & 68.2
        \\
        PointNet++~\cite{qi2017pointnet++}
        & 1,024
        &
        & 77.9
        \\
        DGCNN~\cite{wang2019dynamic}
        &1,024
        &
        &78.1
        \\
        PointCNN~\cite{li2018pointcnn}
        &1,024
        &
        &78.5
        \\
        BGA-PN++~\cite{uy2019revisiting}
        &1,024
        &
        &80.2
        \\
        BGA-DGCNN~\cite{uy2019revisiting}
        &1,024
        &
        &79.7
        \\
        SimpleView~\cite{goyal2021revisiting}
        & 1,024
        &
        & 80.5
        \\
        DynamicScale~\cite{sheshappanavar2021dynamic}
        & 1,024
        &
        & 82.0
        \\
        % Ours*
        % & 1,024
        % & 82.5
        % \\
        \textbf{Ours}
        & 1,024
        &
        & \textbf{85.8 (85.77$\pm$0.06)}
        \\
        \hline
        MVTN~\cite{hamdi2021mvtn}
        & 2,048
        &
        & 82.8
        \\
        \textbf{Ours}
        & 2,048
        &
        & \textbf{87.0 (86.83$\pm$0.16)}
        \\
        \hline
    \end{tabular}
    \label{tab: result_cls}
\end{table}
% \vspace{-9pt}
\subsection{Part Segmentation}
\label{sec: part_seg}
\begin{table}
    \small
    \centering
    \caption{Results of part segmentation on ShapeNet and scene segmentation on S3DIS (Area 5). The three-run best, average, and standard deviations are reported. Note that we report preprocessing methods for the S3DIS dataset because of their significant impact on the final performance~\cite{yan2020pointasnl,xu2021paconv}} 
    \label{tab: result_seg}
    \begin{tabular}{l|c|cc}
        \hline
        Method
        % & Cls. mIoU
        & ShapeNet (I. mIoU)
        & preproc.
        & S3DIS (mIoU)
        \\\hline
        PointConv~\cite{wu2019pointconv}
        % &82.8
        &85.7
        & -
        & -
        \\
        RS-CNN~\cite{liu2019relation}
        % & 84.0
        & 86.2
        & -
        & -
        \\
        CurveNet~\cite{xiang2021walk}
        % & - 
        & 86.8
        & -
        & -
        \\
        AGCN~\cite{kim2021agcn}
        % &83.7
        & \textbf{87.9}
        & -
        & -
        \\
        KPConv-rigid~\cite{thomas2019kpconv}
        % & \textbf{85.1}
        & 86.2
        & Grid
        & 65.4
        \\
        KPConv-d~\cite{thomas2019kpconv}
        % & \textbf{85.1}
        & 86.4
        & Grid
        & 67.1
        \\
        Point Transformer~\cite{zhao2021pointtransformer}
        % &83.7
        & 86.6
        & Grid
        &\textbf{70.4}
        \\
        PointNet~\cite{qi2017pointnet}
        % & 80.4
        & 83.7
        & BLK
        & 41.1
        \\
        PointNet++~\cite{qi2017pointnet++}
        % &81.9
        &85.1
        &BLK
        & 57.3
        \\
        PointCNN~\cite{li2018pointcnn}
        % &84.6
        &86.1
        &BLK
        & 57.3
        \\
        PAConv~\cite{xu2021paconv}
        % &84.6
        &86.1
        & BLK
        & 66.6
        \\
        \hline
        % Ours
        % % & 
        % & 86.8 (86.75$\pm$0.06)
        % \\
        Ours
        % &
        & \underline{87.0} (86.94$\pm$0.08)
        & BLK
        & 66.8(66.72$\pm$0.07)
        \\\hline
    \end{tabular}
\end{table}
We use 2,048 points with normal information as the input. Random anisotropic scaling and random translation are used for data augmentation. SGD is used for optimization. The initial learning rate is set to 0.1. We use voting for post-processing, as it is a common practice~\cite{thomas2019kpconv,liu2019relation,xu2021paconv}. The instance-wise average intersection over union (I. mIoU)~\cite{qi2017pointnet++} is used for the performance assessment. 

\begin{figure}[t]
     \centering
    \includegraphics[width=0.9\linewidth]{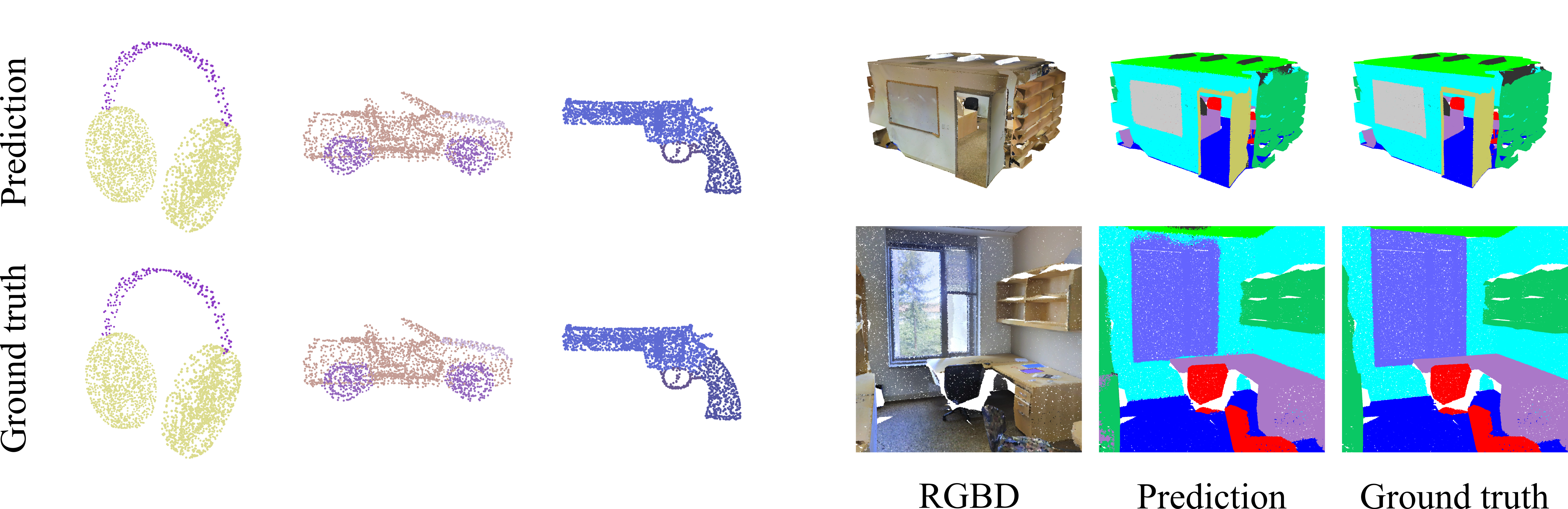}
    \caption{Qualitative results of part and scene segmentation.}
    \label{fig: quali_seg}
\end{figure}

% \vspace{-9pt}
% \paragraph{Result.} 
% \vspace{3pt}
% \noindent\textbf{Result.} 
The results are listed in Table~\ref{tab: result_seg}. The DU-Net achieves competitive performance among cutting-edge models. We believe that DU-Net is especially effective in recognizing object part boundaries, as DUs try to preserve structures while simplifying/smoothing within boundary regions. Although AGCN~\cite{kim2021agcn} achieves strong performance, it relies on a different training setting and a discriminator network with adversarial training in addition to the segmentation network, which considerably increases the complexity. The qualitative results are shown in Fig.~\ref{fig: quali_seg}.
\subsection{Scene Segmentation}
\label{sec: scene_seg}
Similar to \cite{tchapmi2017segcloud}, we advocate using Area five for testing and others for training. Following~\cite{zhao2019pointweb}, we randomly extract a 1m$\times$1m pillar and take 4,096 points as the input. During the test, we test on all points. We use 3D coordinates, RGB, and normalized 3D coordinates with respect to the maximum coordinates in a room as the inputs. We use SGD with an initial learning rate of 0.1 and train the models for 100 epochs, with each epoch set to 1.5k iterations. Random vertical rotation, random anisotropic scaling, Gaussian jittering, and random color dropout are used for data augmentation. The performance is assessed using the point-average IoU. 

% (mIoU), overall accuracy (OA), and class-averaged accuracy (mAcc). 
% \vspace{-5pt}

% \paragraph{Result.}
The results are reported in Table~\ref{tab: result_seg}. As mentioned by \cite{yan2020pointasnl} and \cite{xu2021paconv}, the choice of preprocessing methods has a significant impact on the results, which makes comparisons of methods that use different preprocessing procedure difficult.  Primarily, we compare our model with other models using the same preprocessing (BLK). We achieve the best performance under the same preprocessing method. The qualitative results are shown in Fig.~\ref{fig: quali_seg}.

\section{Design Analysis}
\label{sec: design}
\begin{table}
    \small
    \centering
    \parbox{.45\linewidth}{
        \centering
        \caption{Results of the ablation study on DU
        % The performance of model D is slightly different from the results reported in Table~\ref{tab: result_seg} because the voting post-processing is not performed
        }
        \begin{tabular}{lccccc}
        \hline
        Model
        & \#DU
        &$\phi$
        &$\varphi$
        & \#neigh.
        & I. mIoU
        \\\hline
        
        A
        &1
        &
        & 
        & 16
        & 85.4
        \\
        B
        &1
        & \checkmark
        &
        & 16
        & 86.3
        \\
        C
        &1
        & 
        & \checkmark
        & 16
        & 86.3
        \\
        D
        &1
        & \checkmark
        & \checkmark
        & 16
        & \textbf{86.8}
        \\
        E
        & 2
        & \checkmark
        & \checkmark
        & 16
        & 86.7
        \\
        F
        & 3
        & \checkmark
        & \checkmark
        & 16
        & \textbf{86.8}
        \\
        G
        &1
        & \checkmark
        & \checkmark
        & 4
        & 86.5
        \\
        H
        &1
        & \checkmark
        & \checkmark
        & 8
        & 86.6
        \\
        I
        &1
        & \checkmark
        & \checkmark
        & 24
        & 86.7
        \\\hline
        % mIoU
        % & 86.8
        % & 86.3
        % & 86.3
        % & 85.4
        % & 86.4
        % &
        % \\
        
    \end{tabular}
    \label{tab: ablation_DU}
    }
    \hfill
    \parbox{.48\linewidth}{
    \centering
        % \caption{Comparing various point convolutions. We use our model without DUs as a common backbone. LightPConv (DSC) denotes the LightPConv that adopts the standard DSC scheme. Speed indicates running time per batch (ms)}
    %     \begin{tabular}
    %     % {lcc}
    %     {lccc}
    %     \hline
    %     Method
    %     &I. mIoU
    %     &\#param
    %     % &Latency (ms)
    %     &Speed
    %     \\\hline
    %     % \midrule
    %     PointConv~\cite{wu2019pointconv}
    %     & 86.0 
    %     & 18.97
    %     & 18.9
    %     \\
    %     RSConv~\cite{liu2019relation}
    %     & 86.1
    %     & 3.03
    %     & \textbf{17.4}
    %     \\\hline
    %     % \midrule
    %     % KPConv (rigid)~\cite{thomas2019kpconv}
    %     KPConv~\cite{thomas2019kpconv}
    %     & 86.0
    %     & 10.18
    %     & 19.5
    %     \\
    %     LightPConv
    %     & \textbf{86.4}
    %     & \textbf{3.01}
    %     & \underline{18.5}
    %     \\
    %     LightPConv (DSC)
    %     & 86.0
    %     & 3.50
    %     & 20.0
    %     % \\
    %     % DALC
    %     % & \textbf{86.8}
    %     % & 4.00
    %     % % & 18.5
    %     \\\hline
    %     % \bottomrule
    % \end{tabular}
    \caption{
        Incorporating DUs with various convolutions.
        % We use the models without DUs as the backbone. 
        We take the architecture in Fig.\ref{fig: model} as the base model and investigate the performance change by replacing KPConv-l with other convolution methods or unstacking DUs.
        The reported performance indicate I. mIoU
    }
    \begin{tabular}
        % {lcc}
        {lccc}
        \hline
        Convolutions
        &w/o DU
        &w/ DU
        % &Latency (ms)
        &$\Updelta$
        \\\hline
        % \midrule
        PointConv~\cite{wu2019pointconv}
        & 86.0
        & 86.7
        & \textbf{+0.7}
        \\
        RSConv~\cite{liu2019relation}
        & 86.1
        & 86.4
        & +0.3
        \\
        % \midrule
        % KPConv (rigid)~\cite{thomas2019kpconv}
        KPConv~\cite{thomas2019kpconv}
        & 86.0
        & 86.3
        & +0.3
        \\
        % KPConv-DSC
        % & 86.0
        % & 86.1
        % & +0.1
        % \\
        \hline
        KPConv-l (ours)
        & \textbf{86.4}
        & \textbf{86.8}
        & +0.4
        \\\hline
        % \bottomrule
    \end{tabular}
    \label{tab: compare_conv}
    }
    
\end{table}

\begin{table}[]
    \centering
    \caption{Comparison of DU with other edge-aware methods. We take the architecture in Fig.\ref{fig: model} as the base model and replace DUs other methods }
    \begin{tabular}{lccc}
        \hline
         Method
         & I. mIoU
         & \#params (M)
         & Running time (ms)
         \\
         \hline
         EdgeConv~\cite{wang2019dynamic}
         & 85.5
         & \textbf{4.0} % 4.002639
         & 28.4
         \\
         Point Trans.~\cite{zhao2021pointtransformer}
         & 86.5
         & 5.6
         & 27.9
         \\
         \hline
         DU
         & \textbf{86.8}
         & \textbf{4.0}
         & \textbf{26.6}
         \\
         \hline
    \end{tabular}
    
    \label{tab:comparison_edge_aware}
\end{table}

\begin{table}[]
    \centering
    \caption{Comparison of KPConv and KPConv-l in terms of memory consumption and inference time with different numbers of input points}
    \begin{tabular}{llccccc}
        \hline
         
         & Method 
         & 10k
         & 20k
         & 40k
         & 80k
         \\
         \hline
         \multirow{2}{*}{Memory (M)}
         & KPConv~\cite{thomas2019kpconv}
         & 1894
         & 2246
         & 3068
         & 4616
         \\

         & KPConv-l
         & 1608 
         & 1844
         & 2288
         & 3096
         \\\hline
         \multirow{2}{*}{Inference (ms)}
         & KPConv~\cite{thomas2019kpconv}
         & 94.5
         & 300.5
         & 952.7
         & 3491.9
         \\

         & KPConv-l
         & 88.4
         & 275.6
         & 922.0
         & 3440.9
         \\\hline
    \end{tabular}
    
    \label{tab:comparison_memory_time}
\end{table}

In this section, we validate the design choices regarding DU. All experiments are conducted on the ShapeNet (without voting post-processing) because the part segmentation task is sufficiently complex. 

% \subsection{Design analysis}
\vspace{3pt}
\noindent\textbf{DU components.}
We investigate the influence of DU components. The results are listed in Table~\ref{tab: ablation_DU} (A, B, C, and D). Removing both $\phi$ and $\varphi$ leads to significantly degraded performance. Equipping DU with only $\phi$ or $\varphi$ achieves acceptable performance (models B and C). The performance reaches a peak when both components are incorporated into DU (model F), which successfully verifies our design choice.  

% \vspace{-9pt}
% \paragraph{\#DU.} 
\vspace{3pt}
\noindent\textbf{Number of DUs.}
We investigate the impact of the number of DUs applied to each KPConv-l. 
% Specifically, each DU in the original model is replaced by a series of DUs. 
As shown in Table~\ref{tab: ablation_DU} (D, E, and F), the performance is not sensitive to the number. Consequently, we set \#DUs=1 for all tasks. 
% \vspace{-9pt}

\vspace{3pt}
\noindent\textbf{Neighborhood size.}
% \paragraph{Neighborhood size.} 
As listed in Table~\ref{tab: ablation_DU} (D, G, H, and I), the best performance is achieved when \#neigh is selected as 16; therefore, we use 16 as our default choice. 
% \vspace{-9pt}

% \vspace{3pt}
% \noindent\textbf{Comparing various point convolutions.}
% % \paragraph{Comparing various point convolutions.}  
% The results are listed in Table~\ref{tab: compare_conv}. Compared with KPConv, LightPConv outperforms it concerning the performance and computational efficiency simultaneously. Moreover, LightPConv outperforms LightPConv*, which verifies the effectiveness of the proposed inverse depthwise separable scheme over the standard one. LightPConv also achieves the best performance and parameter-efficiency while achieving competitive running time efficiency among all tested convolutions. 

\vspace{3pt}
\noindent\textbf{Integrating DU with various convolutions.} To show the general applicability of DU, we replace KPConv-l with various popular point convolutions and perform a comparative study. As shown in Table~\ref{tab: compare_conv}, DU successfully boosts other convolution methods, demonstrating its general applicability. 

\vspace{3pt}
\noindent\textbf{DU \textit{vs.} other edge-aware methods.} We compare DU with EdgeConv~\cite{wang2019dynamic,li2019deepgcns,pan2020ecg} and Point Transformer~\cite{zhao2021pointtransformer}. For a fair comparison, we replace DUs with other methods in our model. The result is listed in Table~\ref{tab:comparison_edge_aware}. Evidently, DU is superior to or on par with other methods in terms of performance, number of parameters, and running time.

\vspace{3pt}
\noindent\textbf{KPConv-l \textit{vs.} KPConv.} As shown in Table~\ref{tab:comparison_memory_time}, KPConv-l has a faster inference speed and requires less memory consumption than KPConv. This indicates the usefulness of our trick in the design of KPConv-l.

\section{Conclusion}
Learning point clouds are difficult due to the lack of connectivity information (i.e., edges). Various edge-aware methods in which constructed edges are used as additional information are proposed, which successfully improve the performance. However, \textit{how} the edges improve the performance is hardly interpretable. In this study, we go beyond edge awareness, and propose the diffusion unit (DU) that performs edge enhancement and suppression adaptively in an interpretable manner. A theoretical analysis is conducted to reveal the underlying mechanism of DU, which is confirmed by the following qualitative analysis using the smoothness information. Concretely, the above analysis reveals that DU learns to enhance task-related edges while suppressing others. Extensive experiments show that the network powered by DUs (DU-Nets) can achieve competitive performance across various challenging benchmarks. In particular, DU-Net achieves the state-of-the-art performance in point cloud classification. 
\bibliographystyle{splncs04}
\bibliography{egbib}
\end{document}